\documentclass[letterpaper]{article} 
\usepackage[utf8]{inputenc}
\usepackage[margin=1in]{geometry}

\usepackage{graphicx}  
\usepackage{times}  
\usepackage[hyphens]{url}
\usepackage{graphicx}
\usepackage{amsmath}
\usepackage{tabularx, booktabs}
\usepackage{multirow}
\usepackage{bbm}
\usepackage{bm}
\usepackage{authblk}

\title{On the relationship between multitask neural networks and multitask Gaussian Processes}

\author[1]{Karthikeyan K}
\author[2]{Shubham Kumar Bharti\thanks{This work was done while at IIT Kanpur} }
\author[1]{Piyush Rai}
\affil[1]{Department of Computer Science and Engineering, Indian Institute of Technology Kanpur}
\affil[2]{Department of Computer Science, University of Wisconsin-Madison}
\affil[1]{\{kkarthi, piyush\}@cse.iitk.ac.in}
\affil[2]{\{sbharti\}@wisc.edu}

\begin{document}

\maketitle

\begin{abstract}
Despite the effectiveness of multitask deep neural network (MTDNN), there is a limited theoretical understanding on how the information is shared across different tasks in MTDNN. In this work, we establish a formal connection between MTDNN with infinitely-wide hidden layers and multitask Gaussian Process (GP). We derive multitask GP kernels corresponding to both single-layer and deep multitask Bayesian neural networks (MTBNN) and show that information among different tasks is shared primarily due to correlation across last layer weights of MTBNN and shared hyper-parameters, which is contrary to the popular hypothesis that information is shared because of shared intermediate layer weights. Our construction enables using multitask GP to perform efficient Bayesian inference for the equivalent MTDNN with infinitely-wide hidden layers. Prior work on the connection between deep neural networks and GP for single task settings can be seen as special cases of our construction. We also present an adaptive multitask neural network architecture that corresponds to a multitask GP with more flexible kernels, such as Linear Model of Coregionalization (LMC) and Cross-Coregionalization (CC) kernels.  We provide experimental results to further illustrate these ideas on synthetic and real datasets. 
\end{abstract}

\section{Introduction}
\label{intro}


Multitask learning (MTL) is a learning paradigm in which multiple tasks are learned jointly, aiming to improve the performance of individual tasks by sharing information across tasks~\cite{caruana1997multitask,mtlsurvey}, using various information sharing mechanisms. For example, MTL models based on deep neural networks commonly use shared hidden layers for all the tasks; probabilistic MTL models are usually based on shared priors over the parameters of the multiple tasks~\cite{passos2012flexible,chelba2006adaptation}; Gaussian Process based models, e.g., multitask Gaussian Processes (GP) and extensions~\cite{bonilla2008multi,williams2009multi}, commonly employ covariance functions that models both inputs and task similarity. Multi-label, multi-class, multi-output learning can be seen as special cases of multitask learning where each task has the same set of inputs. 

Transfer learning is also similar to MTL, except that the objective of MTL is to improve the performance over all the tasks whereas the objective of transfer learning is to usually improve the performance of a target task by leveraging information from source tasks ~\cite{mtlsurvey}. Zero-shot learning and few-shot learning are also closely related to MTL. 

Prior works ~\cite{neal1996priors,williams1997computing} have shown that a fully connected Bayesian neural network (NN)~\cite{neal1993bayesian,neal2012bayesian} with a single, infinitely-wide hidden layer, with independent and identically distributed (i.i.d) priors on weights, is equivalent to a Gaussian Process. The result has recently been also generalized to deep Bayesian neural networks \cite{lee2017deep} with any number of hidden layers. These connections between Bayesian neural networks and GP offer many benefits, such as theoretical understanding of neural networks, efficient Bayesian inference for deep NN by learning the equivalent GP, etc.

Motivated by the equivalence of deep Bayesian neural networks and GP, in this work, we investigate whether a similar connection exists between deep \emph{multitask} Bayesian neural networks~\cite{ruder2017overview} and \emph{multitask} Gaussian Processes ~\cite{bonilla2008multi}. Our analysis shows that, for multitask Bayesian NN with single as well as multiple hidden layers, there exists an equivalent multitask GP (under certain priors on neural network weights). Furthermore, we derive the multitask GP kernel function corresponding to multitask Bayesian NN with single as well as multiple hidden layers.

By leveraging this connection, we show that information among multiple tasks is shared due to shared priors on weights of neural network (which corresponds to kernel hyperparameters in GP) and correlation between weights from last hidden layer to output layer (corresponds to task correlations of multitask GP), which is contrary to the common belief that information among different tasks is shared because of shared hidden layers of the neural network. 

Our analysis shows that simple hard parameter sharing ~\cite{ruder2017overview} multitask NN corresponds to multitask GP with the Intrinsic Coregionalization Model (ICM) kernel ~\cite{alvarez2012kernels}. Further exploiting the MTDNN and multitask GP connection, we design a novel and more flexible adaptive multitask NN architecture that corresponds to multitask GP with Linear Model of Coregionalization (LMC) and Cross-Coregionalization (CC) kernels ~\cite{alvarez2012kernels}. To summarize, our contributions are as follows: 

\begin{itemize}
    \item We establish a formal connection between multitask Bayesian NN and multitask GP. Furthermore, we derive the kernel function for the corresponding multitask GP which turns out to be an ICM kernel ~\cite{alvarez2012kernels}. 
    \item We provide a better theoretical understanding on how information is shared in multitask Bayesian NN, in particular, we show that information among multiple tasks is shared due to shared priors on weights of neural network and correlation between weights from last hidden layer to the output layer.
    \item We propose a novel and more flexible multitask neural network architecture which we call  'Adaptive multitask neural network (AMTNN)' that corresponds to multitask GP with more flexible LMC and CC kernels.
\end{itemize}

\section{Notations}

\begin{figure}
\centering
    \fbox{
    \includegraphics[scale=0.4]{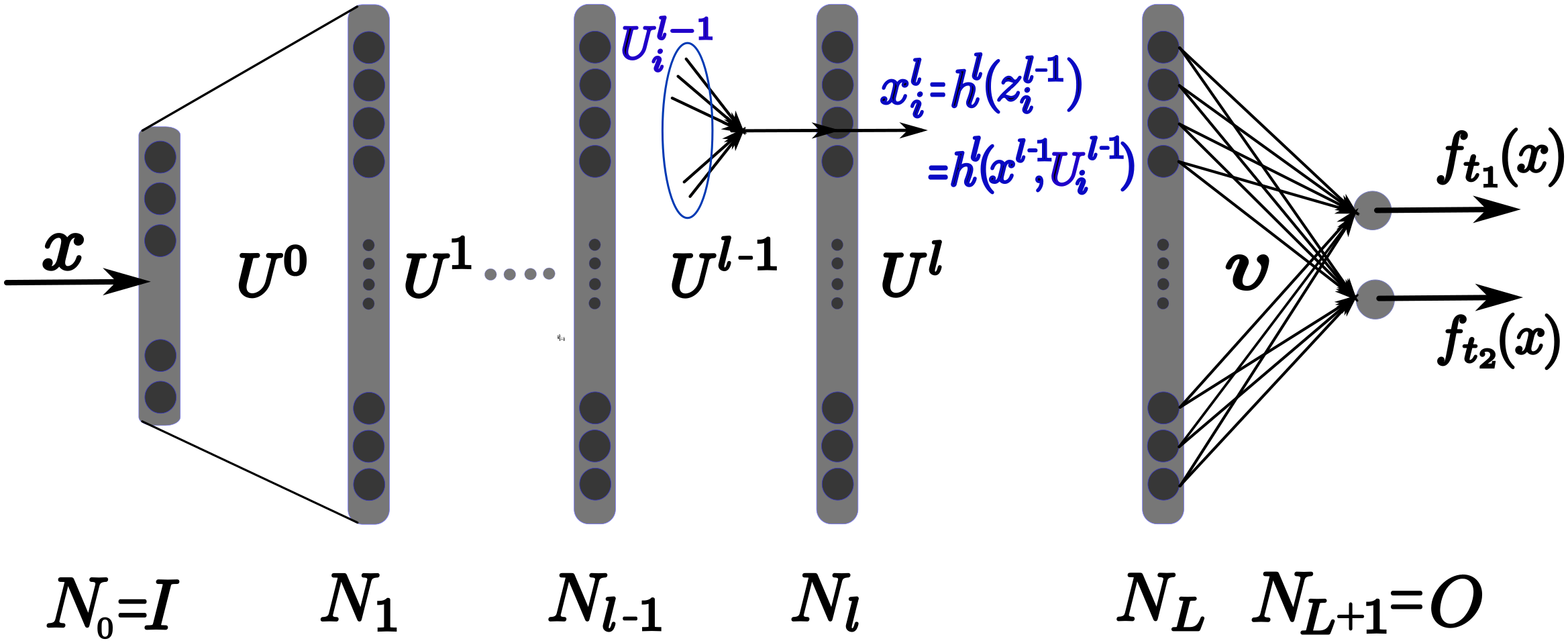}}
    \caption{A multi-task deep neural network with 2 tasks}
    \label{fig:mtnn}
\end{figure}

\label{notation}
Let us consider a deep neural network with $L$ hidden layers, with $l^{th}$ layer having $N^l$ number of hidden units. We denote the weights from  $(l-1)^{th}$ layer to $l^{th}$ layer as $U^{l-1}$ (matrix of size $N^{l} \times N^{l-1}$) and output of $l^{th}$ layer by $x^{l}$ (vector of size $N^{l}$). The bias for $i^{th}$  node in a $l^{th}$ layer is denoted by $b^{l-1}_i$. 

Let us denote the input to $i^{th}$ node of $l^{th}$ layer by $z^{l-1}_i$. Therefore its output $x^{l}_{i}$ can be expressed as 
\[
    x^l_i = h^l(z^{l-1}_i) = h^l\Big(\sum_{j=1}^{N^{l-1}} U^{l-1}_{ij}x^{l-1}_j+b^{l-1}_i\Big)
\]
where $h^l$ is the activation function of all nodes in $l^{th}$ layer. For conciseness, we often write it as 
\begin{align*}
x^l_i &= h^l(z^{l-1}_i) = h^l( x^{l-1},U^{l-1}_{i}) \\
x^l   &= h^l(z^{l-1}) = h^l(x^{l-1},U^{l-1})
\end{align*}
Following the notation in ~\cite{lee2017deep}, we denote the first layer input as $x$. We then have 
\begin{align*}
x^l(x) = h^l(z^{l-1}(x)) = h^l(x^{l-1}(x),U^{l-1})
\end{align*}
Just to distinguish the last (output) layer from all other layers of the neural network, we will denote the last layer weights $U^L$ as $v$. Also, we denote input layer $x^0(x)$ as $x$ and output layer $x^{L+1}$ as $f(x)$. The last layer has identity transfer function $h^{L+1}(\alpha) = \alpha$. Using this convention, the final layer output can be expressed as 
\begin{align}
f(x) = x^{L+1}(x) = h^{L+1}(z^{L}(x)) = z^{L}(x)
\end{align}
We often refer single hidden layer Neural network as single-layer neural network and number of nodes in a layer as the width of the layer.

\section{Neural Networks and Gaussian Processes}\label{sec:related}
In this section, we briefly discuss prior works on the connection between single and deep Bayesian NN with GP. 

\subsection{Single-Layer Neural Network as Gaussian Process}
Neal~\cite{neal1996priors} defined priors over weight and biases of single-layer Bayesian NN and showed that, in the limit of infinite width, the neural network converges to a GP. In a single-layer Neural network with $N^1=H$ hidden units, the output $f(x)$ (single task) can be expressed as
\begin{align}
    f(x) &= b + \sum_{j=1}^{H} v_j h(x , U_j)
\end{align}
Here, $b$, $v$ and $U_j$'s are assumed to be mutually independent. $v_j$'s and $U_j$'s are i.i.d. with Gaussian prior $\mathcal{N}(0, \sigma_v^2)$ and $b$ has a Gaussian prior $\mathcal{N}(0,\sigma_b^2)$. 

\subsection{Deep Neural Networks as Gaussian Processes}
Hazan and Jaakkola \cite{hazan2015step} extended Neal's~\cite{neal1996priors} work to a two hidden layer neural network and further derived the stochastic kernel function for the corresponding GP. Lee et al \cite{lee2017deep} further extended this connection to deep networks. The connection was based on exploiting the independence of the outputs of hidden layer nodes and applying the central limit theorem recursively on the intermediate layers. 

For a deep neural network, suppose $K^1$ denotes the kernel based on the first hidden layer's output
\begin{align*}
    K^1(x_1,x_2) = E \big[z_i^1(x_1) z_i^1(x_2)\big] 
\end{align*}
~\cite{lee2017deep} showed that as $N_l \rightarrow\infty \quad \forall l\in[L]$, a \emph{deep} neural network converges to a GP; $f \sim \mathcal{GP}(\mu,K)$ with mean $\mu = 0$ and covariance function $K^L$,
\begin{align}
&K^L(x_1,x_2) = E\big[z_i^L(x_1)z_i^L(x_2)\big] \nonumber\\
&=\sigma_b^2+\omega_v^2E_{z_i^{L-1}\sim \mathcal{GP}(0,K^{L-1})} \big[x_i^{L}(x_1)x_i^{L}(x_2)\big] \nonumber\\
&=\sigma_b^2+\omega_v^2 \mathcal{F}_{h}\big(K^{L-1}(x_1,x_1), K^{L-1}(x_1,x_2), K^{L-1}(x_2,x_2)\big)
\end{align}
where the recursive kernel equation turns out to be
\begin{align}
K^l(x_1,x_2)=\sigma_b^2+ \omega_v^2 \mathcal{F}_{h}\big(K^{l-1}(x_1,x_1), K^{l-1}(x_1,x_2), K^{l-1}(x_2,x_2)\big)
\end{align}
 with the base case being $K^0(x_1,x_2) =E\big[z_i^0(x_1)z_i^0(x_2)\big]$ and $\omega^2_{v} = H\sigma^2_{v}$. The functional form of $\mathcal{F}_{h}$ depends on activation function $h$ and some other hyperparameters.

\section{Multitask Neural Network}\label{mtnn-des}
We use fully-connected hard parameter sharing multitask Bayesian neural network architecture~\cite{ruder2017overview,caruana1997multitask} where all the tasks share same weights from input layer till the last hidden layer, and each task has separate weights from last hidden layer to output layer (cf., Fig~\ref{fig:mtnn}). Note that we will be using the \textit{Bayesian} version of the this multitask NN. Throughout this exposition, multitask neural network refers to this hard-parameter sharing multitask Bayesian Neural Network (unless specified  otherwise). Later, in Section~\ref{sec:adaMTNN}, we design a new neural network architecture - an adaptive multitask Bayesian neural network.

\section{Multitask Gaussian Process}
\label{mtgp-des}
Multitask GP (MTGP)~\cite{bonilla2008multi} defines a joint prior distribution over multiple functions, where each function models a learning task (e.g., regression or classification). Let there be $T$ tasks and let $f_1, f_2, ... f_T$ denote the $T$ functions such that the $k^{th}$ task is modelled by  $f_k$. Let $X=[x_1, x_2, ...x_N]$ be $N$ input data points with corresponding tasks  $\mathcal{T}=[t_1, t_2, ...t_N], t_i \in [T] \quad \forall i$, and observed scalar outputs $y = [y_1, y_2, ...y_N]$, i.e. $y_i$ is the observed output of $t_i^{th}$ task on input $x_i$. Given $f_{t_i}$, the likelihood of $y_i$ is expressed as 
\begin{align*}
     y_i | f_{t_i}(x_i) &\sim \mathcal{N}(f_{t_i}(x_i), \sigma^2_{t_i})
\end{align*}
where $\sigma^2_{t_i}$ is the noise variance of $t_i^{th}$ task. 

Let $f$ denote the joint function over all tasks, i.e $f$ takes data point $x_i$ and its task $t_i$ as input and outputs $f_{t_i}(x_i)$
\begin{align*}
    f(x_i, t_i) &= f_{t_i}(x_i)
\end{align*}
Multitask GP defines a GP prior on joint function $f$, i.e. any finite sample of $f(x, t)$ forms a Gaussian distribution
\begin{align*}
\begin{bmatrix}
        f(x_1, t_1)\\
        f(x_2, t_2)\\
        .\\
        .\\
        f(x_s, t_s)
    \end{bmatrix} 
    &=        \begin{bmatrix}
        f_{t_1}(x_1)\\
        f_{t_2}(x_2)\\
        .\\
        .\\
        f_{t_s}(x_s)
    \end{bmatrix}
    \sim 
    \mathcal{N}
    \begin{pmatrix}
        \begin{bmatrix}
            \mu_{t_1}(x_1)\\
            \mu_{t_2}(x_2)\\
            .\\
            .\\
            \mu_{t_s}(x_s)
        \end{bmatrix}, 
    \Sigma\\
    \end{pmatrix}\\
    \Sigma[i, j] &= Cov(f_{t_i}(x_i) , f_{t_j}(x_j))\\
    &= K(x_i, t_i, x_j, t_j)
\end{align*}
where $\{x_i, t_i\}$  are any random set of input and task samples. $K(x_i, t_i, x_j, t_j)$ is a task dependent kernel function and $\mu_{t}(x)$ is the mean function which is usually assumed to be zero. For simplicity, it is usually assumed that task dependent kernel can be decomposed into two parts - covariance between tasks and covariance between inputs~\cite{bonilla2008multi,alvarez2012kernels}.
\begin{align}
    K(x_i, t_i, x_j, t_j) = K_{task}(t_i, t_j)K_{input}(x_i, x_j)
\end{align}
With the above-defined priors and likelihood, the posterior predictive distribution turns out to be Gaussian. Please refer the appendix \textit{Multitask Gaussian Process} for derivations of posterior predictive and further discussions on how information is shared in a multitask GP.

\section{Multitask Bayesian NN as Multitask GP} \label{sec:mtbnn-mtgp}
In this section, we define priors over weights and biases of \emph{single-layer} multitask Bayesian NN and show that it converges to multitask GP in the limit of infinite width. Next, we derive the corresponding MTGP kernel function. At the end of this section, we also present some important and surprising observations based on this connection.

\subsection{Priors on single-layer multitask Bayesian NN}

In case of multi-output learning, Neal~\cite{neal1996priors} assumed outputs to be not related and stated that training an infinitely wide multi-output neural network is the same as training for each output separately. In contrast, we assume outputs of different tasks to be potentially related (note that task relatedness is central to multitask learning). 

Let us consider a single hidden layer multitask neural network with $T$ tasks and $H$ hidden units, with the activation function of each hidden units being $h$. Let $U_i$ be the weights from input to $i^{th}$ hidden unit and $v_j^k$ be the weights from $j^{th}$ hidden unit to output of $k^{th}$ task and $b_k$ be the bias from hidden unit to output of $k^{th}$ task. This architecture corresponds to Fig.~\ref{fig:mtnn} with $L = 1 $ and $N_1 = H$. The output of  $k^{th}$ task, $f_k$ can be written as   
\begin{align}
    f_k(x) = b_k + \sum_{j=1}^{H} v_j^k h(x, U_j),\text{ where } x \in \mathbbm{R}^{d} 
\end{align}
We define zero mean i.i.d Gaussian prior on $U_i$'s 
\begin{align*}  
U_i &\sim \mathcal{N}(0, \Sigma_u), \forall i, U_i \in \mathbbm{R}^{d}
\end{align*}
For a given task $k$, we assume $v_j^{k}$ to be i.i.d. with a zero mean Gaussian prior and $\sigma^2_{kk}$ variance.
\begin{align*}
    v_j^{k} \sim \mathcal{N}(0, \sigma^2_{kk}), \forall j
\end{align*}
For $i \neq j$, assume $v_i^{k_a} $ and  $v_j^{k_b} $ are independent, and for a given $j$, assume $v_j$'s \emph{can be correlated  across different tasks}, i.e.  $Cov(v_j^{k_a}, v_j^{k_b})$ need not be zero. We also assume, for a given pair of tasks, covariance remains same for all $j$'s. 
\begin{align*}
    Cov(v_i^{k_a}, v_j^{k_b}) &=  \mathbbm{I}(i = j) \sigma^2_{k_ak_b}, \forall i,j
\end{align*}
We define $H\sigma^2_{k_ak_b} = \omega^2_{k_ak_b}$ i.e. $\sigma^2_{k_ak_b}$ scales with $H$. We can introduce a temporary variable $z$, such that $z_i^{t} = \sqrt{H}v_i^{t} $ then $z_i^{t}$ has a zero mean and following covariance,
\begin{align*}
    Cov(z_i^{k_a}, z_j^{k_b}) &=  \mathbbm{I}(i = j) \omega^2_{k_ak_b}, \forall i,j
\end{align*}
We define prior over $b_k$ to be Gaussian with zero mean and following covariance function,
\begin{align*}
    b_k &\sim N(0, C\omega^2_{kk}) \text{ where  C is a constant} \\
    Cov&(b_{k_i}, b_{k_j} ) = C\omega^2_{k_ik_j}
\end{align*}

\subsection{Single-layer Multitask BNN Converges to MTGP}\label{smtnnTOmtgp}

In this section, we show that, as the number of hidden units tends to infinity, single hidden layer multitask BNN with the above-mentioned priors converges to multitask GP. \\ \\
\textbf{Claim 1 : $f_k(x)$ is Gaussian}
\begin{align}
    f_k(x) &= b_k + \sum_{j=1}^{H} v_j^k h(x, U_j) \\
    f_k(x) &= b_k + \sqrt{H}\Big[ \frac{1}{H}\sum_{j=1}^{H} z_j^k h(x, U_j) \Big]
\end{align}
$U_j$ are independent of $z_j^k\text{ }, \forall k,j$ (since $U_j$ are independent of $v_j^k$) 
\begin{align}
    E[z_j^k h(x, U_j) ] &= E[z_j^k]E[h(x, U_j)] = 0 \\
    Var(z_j^k h(x, U_j )) &=  \omega^2_{kk}E[(h(x, U_j))^2]
\end{align}
Assuming $E[h(x, U_j)^2 ]$ is finite (if $h$ is bounded then it is trivially true, but even for ReLU, it is finite~\cite{cho2009kernel}), the variance of  $z_j^k h(x, U_j) $ is finite. We know that 
\begin{itemize}
    \item $\forall j_1 \neq j_2$ $z_{j_1}^k$ and $z_{j_2}^k$ are i.i.d and $U_{j_1}$ and $U_{j_2}$ are i.i.d.
    \item $\forall j_1$ and $j_2 \text{ } z_{j_1}^k$ and $U_{j_2}$ are independent.
\end{itemize}
Therefore, for a given task $k$, $z_j^kh(x, U_j)$'s are  i.i.d. for different $j$. In the limit of $H\rightarrow\infty$, using Central Limit Theorem, 
$\alpha = \sqrt{H}\Big[ \frac{1}{H}\sum_{j=1}^{H} z_j^k h(x, U_j) \Big]$ converges to a Gaussian distribution. $b_k$ is also Gaussian and is independent from $\alpha$. By the sum of two independent Gaussians property, $f_k(x)$ converges to a Gaussian distribution. \\ \\ 
\textbf{Claim 2 : Multitask Bayesian NN Converges to MTGP}\\
We now show that joint priors on $f_k$'s converge to multitask GP prior, by proving that, for any finite subset of inputs $\{x_1 , x_2, ...x_s \}$ and their corresponding tasks, $\{t_1 , t_2, ...t_s\}$, the outputs of multitask Bayesian NN are jointly Gaussian.
\begin{align}
    \begin{bmatrix}
        f_{ t_1}(x_1)\\
        f_{t_2}(x_2)\\
        .\\
        .\\
        f_{t_s}(x_s)
    \end{bmatrix}
    &=
    \begin{bmatrix}
        b_{t_1}\\
        b_{t_2}\\
        .\\
        .\\
        b_{t_2}
    \end{bmatrix} + 
        \sqrt{H}\Big[\frac{1}{H} \sum_{j=1}^{H} F_j \Big]\\
    \text{where } F_j &= 
       [ z_j^{t_1} h(x_1, U_j),... z_j^{t_s} h(x_s, U_j) ]^T \\
       E(F_j) &= [0, 0,... 0], \forall j \\
       Cov(F_j, F_k)_{l,m} & = \begin{cases}
    0 & \text{if $j \neq k $} \\
    \omega^2_{lm} E[h(x_l, U_j)  h(x_m, U_k)] & \text{if $j=k$} \\
  \end{cases}
\end{align}
If we assume $E[h(x_l, U_j)  h(x_m, U_j)]$ to be finite (if $h$ is bounded then it is trivially true, but even for ReLU, it is finite~\cite{cho2009kernel}) then each of the elements in variance-covariance matrix of $F_j$ is finite. 

Note that $F_j$ and $F_k$ are independent for all $j \neq k$ (since every co-ordinate in $F_j$ is independent of every co-ordinate in $F_k$, as in claim 1). Also, since $\{z_j\}$'s (and $\{U_j\}$'s) are identically distributed for different $j$'s, $F_j$'s are also identically distributed. Therefore, by the \emph{multidimensional} central limit theorem, in limit of $H\rightarrow\infty,  \sqrt{H}\Big[\frac{1}{H} \sum_{j=1}^{H} F_j \Big] $ converge to a multivariate Gaussian distribution. 
Since $[b_{t_1}, b_{t_2} .. b_{t_s}]^T$  is Gaussian,   $[ f_{t_1}(x_1), f_{t_2}(x_2), .. f_{t_s}(x_s)]$ is also Gaussian as sum of two independent Gaussians is also Gaussian. 

Thus multitask neural network priors on the functions $f_k$'s jointly converge to a multitask GP prior, establishing the equivalence between the two.

\subsection{MTGP Kernel for single-layer multitask BNN}
In this section we derive the multitask GP kernel functions corresponding to single-layer multitask Bayesian NN, with the above-mentioned priors (in the limit $H \rightarrow \infty$). Since $f_k(x) = b_k + \sum_{j=1}^{H} v_j^k h(x, U_j)$, it is easy to see that the multitask GP mean function $E[f_k(x)]  = 0$.

We can derive the multitask GP covariance function as follows
\begin{align*}
    &Cov(f_{t_1}(x_1), f_{t_2}(x_2)) = E[f_{t_1}(x_1)f_{t_2}(x_2)]\\
    & = E[b_{t_1}b_{t_2}] + E[(\sum_{j=1}^{H} v_j^{t_1} h(x_1, U_j))(\sum_{k=1}^{H} v_k^{t_2} h(x_2, U_k))]\\
    & =  C\omega^2_{t_1t_2} +\sum_{j=1}^{H} \sum_{k=1}^{H} E[v_j^{t_1}v_k^{t_2}]E[h(x_1, U_j)) h(x_2, U_k))] \\
    & = C\omega^2_{t_1t_2} + \sum_{l=1}^{H} \sigma^2_{t_1t_2} E[h(x_1, U_l)) h(x_2, U_l))] \\
    & = C\omega^2_{t_1t_2} + H\sigma^2_{t_1t_2} E[h(x_1, U_l)) h(x_2, U_l))]  \text{ for any} l \\
    & = \omega^2_{t_1t_2}(C +  E[h(x_1,U) h(x_2,U)] )
\end{align*}
Thus the multitask GP kernel corresponding to multitask Bayesian neural network is
\begin{align}
K_{t_1t_2}(x_1, x_2) &= Cov(f_{t_1}(x_1), f_{t_2}(x_2))\\
&= \omega^2_{t_1t_2}(C +  E[h(x_1,U)) h(x_2,U))] )
\end{align}
We can see that the multitask GP covariance function corresponding to multitask Bayesian NN factorizes into input dependent and task dependent components, which is similar to ICM kernels\cite{alvarez2012kernels,bonilla2008multi}
\begin{align*}
     K_{t_1t_2}(x_1, x_2) &= K_{input}(x_1, x_2)K_{task}(t_1, t_2)\\
     &= \omega^2_{t_1t_2}(C +  E[h(x,U)) h(x,U))] )
\end{align*}
Comparing both the equations,
\begin{align}
    K_{task}(t_1, t_2)  &\propto  \omega^2_{t_1t_2} \\
  K_{input}(x_1, x_2) &\propto  C +  E[h(x_1, U)) h(x_2, U))   
\end{align}
which is also intuitive as we know in single task case
\begin{align}
    K(x_1, x_2)= \sigma^2_b +  \omega^2E[h(x, u)) h(x, u))]
\end{align}
Hence we proved that multitask Bayesian NN with appropriately defined priors as above converges to a multitask Gaussian process with ICM kernels.

\subsection{Some Observations}
Let $v_j^{t_1}$ and $v_j^{t_2}$ are independent (or uncorrelated), then
\begin{align}
    Cov(v_j^{t_1}, v_j^{t_2}) &=  \sigma^2_{t_1t_2} = 0 \text{ for } t_1 \neq t_2 \\
    K_{task}(t_1, t_2) &= 0 \\
    Cov(f_{t_1}(x_1), f_{t_2}(x_2)& = 0, \forall x_1, x_2
\end{align}
From the above equations, we can see that if we assume weights from last hidden layer nodes to outputs of different tasks to be independent (or uncorrelated), then it is equivalent to learning each task separately (assuming we fixed all the hyperparameters - including kernels). Refer to the Appendix (Section \textit{Multitask Gaussian Process}) for further discussions on posterior prediction distribution and information sharing in multitask Gaussian process

We can also see that in ICM kernels all the task share same $K_{input}$ function which means that the kernel hyperparameters (of $K_{input}$ part) are shared across all the tasks. 

In contrast, in a multitask Bayesian NN with correlated weights from last hidden units to the output of different tasks, the information is shared both due to task correlations and shared hyperparameters. If the weights are not correlated then information is shared only due to shared hyperparameters (shared priors on weights of intermediate layers).

\section{Deep Multitask Bayesian  NN as Multitask GP}
In this section, we generalize our result from single-layer  multitask Bayesian NN to \emph{deep} multitask Bayesian NN. Further, we derive the corresponding multitask GP kernels. 

Note that, in a multitask deep neural network with $L > 1$ hidden layers, the output of $k^{th}$ task can be written as
\begin{align}
f_k(x) = b^L_k + \sum_{j=1}^{N_L} v^k_j h(x^{L-1}(x), U^{L-1}_j)
\end{align}
where $x^{L-1}$ are penultimate hidden layer outputs. The priors on weights and biases from input to the last hidden layer are the same as follows. We define the set $\{U^{l}_{jk}\}$ and set $\{b_k^l\}|_{l=1}^{L-1}$ to be independent and identically distributed with $b_k^l \sim \mathcal{N}(0,\sigma_b^2)$. All $\{v^t_k\}$ are independent of $\{U^{l}_{jk}\}, \{b_k^l\}$. However, $\{v^t_k\}$ are correlated amongst themselves with covariance, $Cov(v_k^{t_i}, v_l^{t_j}) =  \mathbbm{I}(k = l) \sigma^2_{t_it_j}, \forall k,l$, and $\{b_k^L\}$ are correlated with covariance, $Cov(b^L_{t_i},b^L_{t_j}) = \sigma^{2(t_i,t_j)}_b$.

\subsection{Deep multitask BNN converges to MTGP}
In this section, we show that $f_k(x)$'s converges to Gaussian distributions and consequently deep multitask BNN converges to multitask GP.

Since $v_j^k$ is independent of $U^l, \text{ } \forall l \in \{0, 1, .. L\}$, $v_j^k$ is independent of $h(x^{L-1}(x), U^{L-1}_j )$. Thus $\forall j, v^k_j h(x^{L-1}(x), U^{L-1}_j )$ are i.i.d. 
Using CLT and sum of independent Gaussians property, $f_k(x)$ is Gaussian.

We can show that joint priors on $f_k$'s converge to a multitask GP following the same arguments as in  Claim 2 of single-layer multitask Bayesian NN converges to MTGP. For more details refer to the Appendix (Section \textit{Deep MTNN converges to Deep MTGP}).

\subsection{MTGP Kernel for \emph{deep} Multitask BNN}
Just like the single hidden layer case, here we derive the MTGP kernel corresponding to the deep multitask BNN. Note that the mean is given by
\begin{align*}
    E[f_{t}(x)] &= E[ b^L_k] + \sum_{j=1}^{N_L} E[v^k_j h(x^{L-1}(x), U^{L-1}_j)] = 0
\end{align*}
and covariance function (refer to the Appendix for derivation)
\begin{align}
    &K^{L(t_1t_2)}(x_1,x_2) = E[f_{t_1}(x_1)f_{t_2}(x_2)]  \nonumber\\
    &= \sigma_b^{2(t_1t_2)}+ \omega_v^{2(t_1t_2)}E_{z^{L-1} \sim \mathcal{GP}(0,K^{L-1})} \big[h^L(z^{L-1}(x_1))h^L(z^{L-1}(x_2))\big] \nonumber\\
    &= \sigma_b^{2(t_1t_2)}+\omega_v^{2(t_1t_2)}\mathcal{F}_{h}\big(K^{L-1}(x_1,x_1), K^{L-1}(x_1,x_2), K^{L-1}(x_2,x_2)\big)
\end{align}
We note that, except the last layer, all other layers are task-independent and so is their covariance function. Following~\cite{lee2017deep}, we also note that for a general $l^{th}$ layer the expectation is taken over $z_i^{l-1}\sim \mathcal{GP}(0,K^{l-1})$ and after integration the function can be recursively expressed $\forall l < L$.
The kernel in this case will be
$K^l(x_1,x_2) = \sigma_b^2 + \omega_v^2 \mathcal{F}_{h}\big(K^{l-1}(x_1,x_1), K^{l-1}(x_1,x_2), K^{l-1}(x_2,x_2)\big)$, with the base case kernel being,
$K^0(x,x') =E\big[z^0(x)z^0(x')\big] = \sigma_b^2+\omega_v^2(x \cdot x')$

\section{Adaptive Multitask Bayesian Neural Networks}\label{sec:adaMTNN}
We have seen that hard-parameter sharing multitask Bayesian NN corresponds to multitask GP with ICM kernels. However, ICM kernels are known to be less flexible than Linear Model of Coregionalization (LMC) and Cross-Coregionalization (CC) kernels~\cite{alvarez2012kernels}. Please refer to the Appendix (Section \textit{Limitations of ICM Kernels}) for a detailed discussion on the limitations of ICM kernels. 

In this section, we design a more flexible \emph{adaptive} multitask Bayesian neural network, and show that it corresponds to multitask GP with more flexible LMC and CC kernels.  

\begin{figure}[!htbp]
    \centering
    \fbox{
    \includegraphics[width=0.2\textwidth]{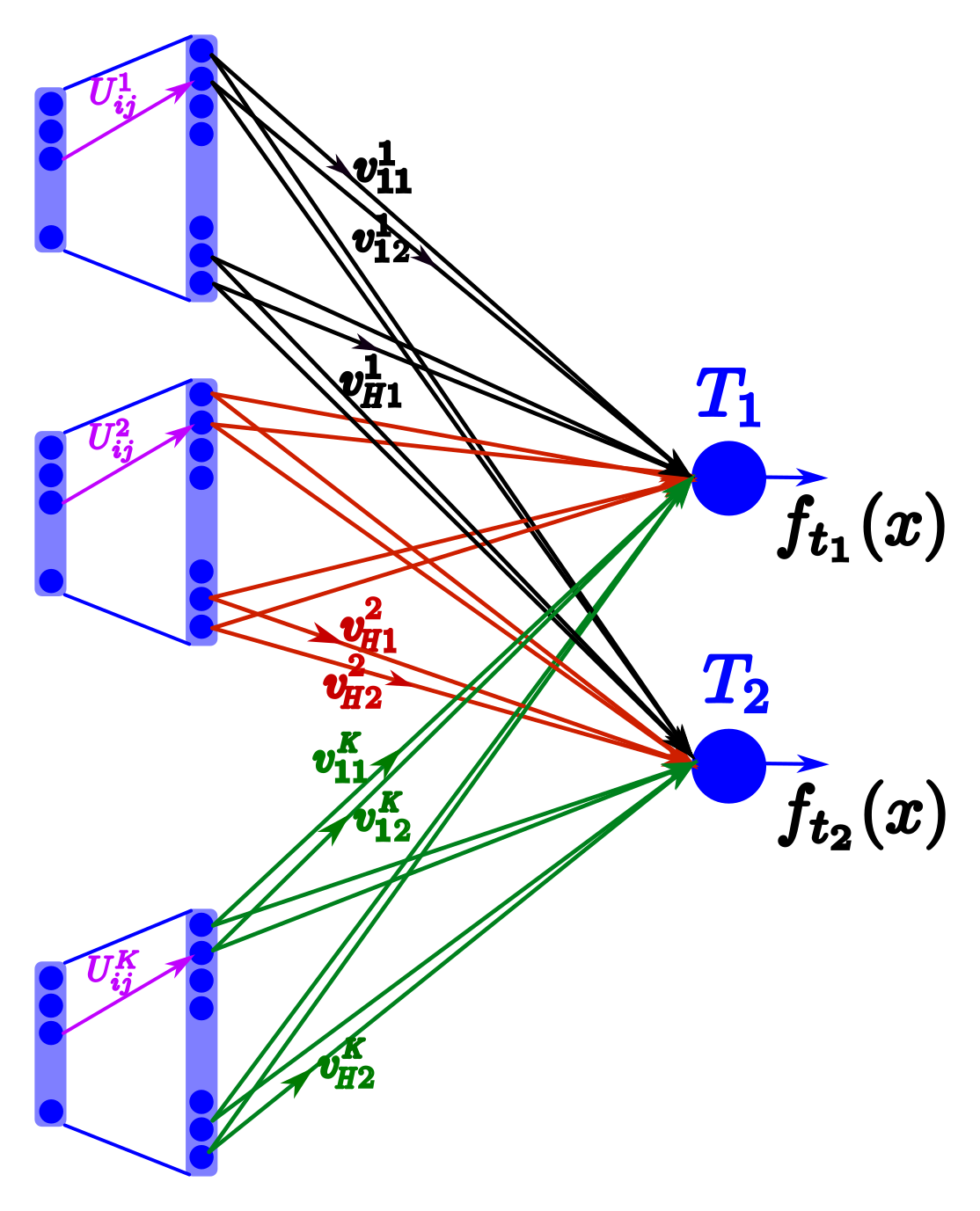}
    }
    \caption{An adaptive multitask neural network with two tasks}
    \label{fig:adap_mtnn}
\end{figure}

\subsection{Adaptive Multitask BNN Architecture and Priors}
Our adaptive multitask BNN architecture has $\mathcal{K}$ basis neural networks (feature extractors). The output of each task is a task-dependent linear combination of features extracted by each of the basis neural networks (Fig.~\ref{fig:adap_mtnn})

For simplicity, we assume all the basis NN to be single hidden layer NNs with $H$ hidden units each. Let $U^{k}_i$ be the weights from input to $i^{th}$ hidden unit of $k^{th}$ basis network. We define zero mean i.i.d Gaussian priors on $U^{k}_i, \text{ } \forall i, k$. Let $v^{k}_{it}$ be the weight from $i^{th}$ hidden unit of $k^{th}$ basis to the output node of $t^{th}$ task. We define a zero mean Gaussian prior on $v^{k}_{it}$ with the following covariance: 
\begin{align*}
    Cov(v^{m}_{it_1} v^{n}_{jt_2} ) = \sigma^{2(mn)}_{t_1t_2} \text{ if } i = j, \text{ and } 0 \text { otherwise }
\end{align*}
We define $U$'s to be independent of the $v$'s. The output of the $t^{th}$ task, $f_t(x)$ can be written as 
\begin{align}
    f_t(x) = \sum_{k=1}^{\mathcal{K}} \sum_{j=1}^{H} v^{k}_{jt} h^{k}(x, U_j^{k})
\end{align}
where $h^{k}$ is the activation function of $k^{th}$ basis NN. For simplicity we do not assume the bias term from hidden unit to output layer.

\subsection{Adaptive MTBNN Converges to Multitask GP}
We show that $f_t$ jointly forms a Gaussian prior 
\begin{align*}
    \begin{bmatrix}
        f_{t_1}(x_1)\\
        f_{t_2}(x_2)\\
        .\\
        .\\
        f_{t_s}(x_s)
    \end{bmatrix}=\sum_{j=1}^{\mathcal{H}} \sum_{k=1}^{K}  F_{jk}= \sum_{j=1}^{\mathcal{H}} \Big(
    \sum_{k=1}^{K} 
        \begin{bmatrix}
            v^{k}_{jt_1} h^{k}(x_1, U_j^{k}) \\
            v^{k}_{jt_2} h^{k}(x_2, U_j^{k})\\
            .\\
            .\\
            v^{k}_{jt_s} h^{k}(x_s, U_j^{k})
        \end{bmatrix}
        \Big)\\
\end{align*}

Note that $G_j = \sum_{k=1}^{K}  F_{jk}$ are i.i.d. with finite variance-covariance.  Therefore, by the multidimensional Central Limit Theorem, $\sum_{j=1}^{\mathcal{H}} G_j $ converges to a Gaussian distribution. Refer to the Appendix (Section \textit{Proof: Adaptive MTBNN Converges to Multitask GP}) for a detailed proof.

\subsection{Adaptive MTGP Kernels }
In this section, we derive the multitask GP kernel corresponding to the adaptive multitask neural network and show that it corresponds to LMC kernels.
First, note that the mean of $f_t(x)$ 
\begin{align*}
    E[f_t(x)] &=  E[\sum_{k=1}^{\mathcal{K}} \sum_{j=1}^{H} v^{k}_{jt} h^{k}(x, U_j^{k} )] = 0
\end{align*}
The covariance function can be derived as follows 
\begin{align*}
    &Cov(f_{t_1}(x_1), f_{t_2}(x_2 ) = E[f_{t_1}(x_1) f_{t_2}(x_2)]\\
    &= E\Big[\Big( \sum_{m=1}^{\mathcal{K}} \sum_{i=1}^{H} v^{m}_{it_1} h^{m}(x_1, U_i^{m}) \Big) \Big( \sum_{n=1}^{\mathcal{K}} \sum_{ j=1}^{H} v^{n}_{jt_2} h^{n}(x_2, U_j^{n}) \Big) \Big] \\
    &= \sum_{m=1}^{\mathcal{K}} \sum_{n=1}^{\mathcal{K}}  \sum_{i=1}^{H} \sum_{ j=1}^{H} E[v^{m}_{it_1} v^{n}_{jt_2} ] E[h^{m}(x_1, U_i^{m}) h^{n}(x_2, U_j^{n})]\\
    &= \sum_{m=1}^{\mathcal{K}} \sum_{n=1}^{\mathcal{K}}  \sum_{i=1}^{H} \sigma^{2(mn)}_{t_1t_2} E[h^{m}(x_1, U_i^{m}) h^{n}(x_2, U_i^{n})]\\
\end{align*}
\begin{align}
    &= \sum_{m=1}^{\mathcal{K}} \sum_{n=1}^{\mathcal{K}}  \omega^{2(mn)}_{t_1t_2} E[h^{m}(x_1, U_l^{m}) h^{n}(x_2, U_l^{n})], \text{ for any } l
\end{align}
Comparing the equation with Cross-Coregionalization (CC) kernels,
\begin{align}
    K_{t_1t_2}(x_1, x_2) &=  \sum_{m=1}^{\mathcal{K}} \sum_{n=1}^{\mathcal{K}} K_{task}^{mn}(t_1, t_2)K_{input}^{mn}(x_1, x_2) \\
    K_{task}^{mn}(t_1, t_2) &\propto \omega^{2(mn)}_{t_1t_2} \\
    K_{input}^{mn}(x_1, x_2)  &\propto E[h^{m}(x_1, U_l^{m}) h^{n}(x_2, U_l^{n})], \text{ for any } l
\end{align}
Therefore, we have shown that adaptive multitask NN converges to multitask GP with CC kernel.

Let us redefine the covariance between $v_{it}^m$ to
\begin{align}
    Cov(v^{m}_{it_1} v^{n}_{jt_2} ) = \sigma^{2(mm)}_{t_1t_2} \text{ if } i=j \text{ and } m=n \text{ and } 0 \text { otherwise }
\end{align}
With the above defined covariance between $v_{it}^m$, 
\begin{align}
    &Cov(f_{t_1}(x_1), f_{t_2}(x_2 ) = \sum_{m=1}^{\mathcal{K}} \omega^{2(mm)}_{t_1t_2} E[h^{m}(x_1, U_l^{m}) h^{m}(x_2, U_l^{n})] 
\end{align}
The above covariance corresponds to an LMC kernel
$ K_{t_1t_2}(x_1, x_2) =  \sum_{m=1}^{\mathcal{K}}  K_{task}^{m}(t_1, t_2)K_{input}^{m}(x_1, x_2)$. 

Therefore, the proposed adaptive multitask NN converges to multitask Gaussian Process with more flexible LMC and CC kernels.

\section{Experiments}
In this section, we present several experiments to illustrate the practical benefits of the connection between multitask Bayesian NN and multitask GP. (Please refer appendix \textit{Experiments} for dataset descriptions) 

\subsection{Efficacy of Multitask BNN vs Multitask GP}
We compare the speed and performance of multitask Bayesian NN (with Variational Bayes (VB) inference) and multitask GP on a subset of SARCOS~\cite{williams2006gaussian} (first 600 and 500 data from train and test, respectively) and Polymer datasets~\cite{xu2013multi,borchani2015survey}. Table~\ref{tbl:mtbnn-vs-mtgp} shows that multitask GP is faster and yields better accuracy than corresponding multitask BNN (for smaller datasets), which shows that VB inference is susceptible to local convergence. Our results also suggests that it is better to use multitask GP for problems like multitask Bayesian Optimization~\cite{swersky2013multi} where often the data size is limited. We use only the first and second outputs (task 1 and task 2) for both SARCOS and Polymer data. We use single layer hard-parameter sharing multitask NN and its corresponding multitask GP.
\begin{table}[htbp]
\centering
\caption{Efficacy of multitask BNN vs Multitask GP}
\begin{tabular}{c  c c c c} 
\toprule
Model &  Epochs & MSE -1 & MSE - 2 & Time (secs)\\
\midrule
\multicolumn{5}{c}{SARCOS - Train 600 and Test 500 Dataset}  \\
\midrule
MTGP  & - & \textbf{48.59} & \textbf{52.06} & 20\\
MTBNN & 10k  & 116.67 & 81.41 & 9.33 \\
MTBNN & 100k   &105.94 & 80.12& 70.69 \\
\midrule
\multicolumn{5}{c}{Polymer Dataset}  \\
\midrule
MTGP  & -& \textbf{0.0037} & \textbf{0.0032} & 0.1\\
MTBNN & 10k  & 0.0179 & 0.0142 & 5.68  \\
MTBNN & 100k  & 0.0128 & 0.0117 & 55.61 \\
\midrule
\bottomrule
\end{tabular}
\label{tbl:mtbnn-vs-mtgp}
\end{table}

\subsection{Correlated vs Uncorrelated Multitask NN}
In this section, we compare the performance of correlated and uncorrelated multitask NN on the Textual-Entailment task - Sentences Involving Compositional Knowledge (SICK) dataset~\cite{sickcure,marelli2014semeval}. We combine SICK data with the Multi-Genre Natural Language Inference (MultiNLI) dataset~\cite{williams2017broad} to make it a multitask problem. (in our experiment task 1 is SICK and task 2 is MultiNLI). As MultiNLI is much bigger than SICK, the performance of MultiNLI is almost unaffected, hence we compare performances only on SICK.

We use standard neural networks rather than Bayesian neural networks, as approximate inference methods often converge to bad optima. For correlated multitask NN, we use correlated regularization on weights from the last hidden layer to the output layer (corresponds to MAP estimate). As we are particularly interested in the importance of correlation between the weights from last hidden layer to output layer, we use Bidirectional Encoder Representations from Transformers (BERT)~\cite{devlin2018bert,liu2019multi} as a feature extractor (i.e we replace dense neural network from input to last hidden layer with BERT-Large). Motivations for using BERT rather than fully connected NN is, 1) its practicality on lots of applications and 2) improving on better models is more useful and reliable than improving on a bad mode (improving performance on simple NN may not be much of use). 

We report best and average accuracy (over 5 runs) on the SICK test set.

\begin{table}[!htbp]
\centering
\caption{Correlated vs Uncorrelated Multitask NN}
\begin{tabular}{ c c c} 
\toprule
\textbf{Model} & \textbf{Best Acc.} & \textbf{Average Acc.} \\
\midrule
\multicolumn{3}{c}{SICK dataset}  \\
\midrule
No Multitask NN & 89.81 & 89.19\\
Uncorrelated multitask NN & 90.93  & 90.78 \\
correlated multitask NN & \textbf{91.23}& \textbf{91.02}\\
\bottomrule
\end{tabular}
\label{tbl:ccr-noccr}
\end{table}

\subsection{Advantage of Adaptive Multitask NN}
In this section, we give a real example where multitask neural networks performs worse than a single task neural network, and we show that adaptive multitask NN performs better than normal multitask neural network.  We use Stanford Sentiment Treebank (SST)~\cite{socher2013recursive} and MultiNLI datasets (SST is task 1 and MultiNLI is task 2) for our experiment and measure classification accuracy on SST. Again, we use BERT for basis neural networks (feature extractor). For adaptive multitask NN, we use two basis. 
Here, we report best and average (over 10 runs) accuracy on SST development data and test accuracy corresponding to the best development model. 

\begin{table}[!htbp]
\centering
\caption{Adaptive vs Normal Multitask NN}
\begin{tabular}{ c c c c} 
\toprule
\textbf{Model} & \textbf{Best Acc.} & \textbf{Average Acc.} & \textbf{Test Acc.}\\
\midrule
\multicolumn{4}{c}{SST Dataset}\\
\midrule
No MTNN & \textbf{92.55}  & \textbf{92.06}  & \textbf{94.12}\\
Normal MTNN &  90.60  &  89.99 & 90.72 \\
Adaptive MTNN &  \textbf{91.63} & \textbf{90.36}  & \textbf{91.98} \\
\bottomrule
\end{tabular}
\label{tbl:adap-mtnn}
\end{table}

From table \ref{tbl:adap-mtnn}, we can see that multitask NN performs much worse than single task NN; this is expected as entailment and sentiment analysis are quite different tasks (also feature vector of Sentiment Analysis corresponds to a single sentence whereas feature vector of entailment corresponds to two sentences). Further, adaptive multitask NN performs better than normal multitask NN. Please refer to appendix for a similar experiment on simulated dataset and fully connected neural networks.

\section{Conclusion and Future Work} 
We have shown that multitask Bayesian deep neural networks converge to a multitask GP and derived its corresponding multitask GP kernels. Our analysis sheds light on the behavior of multitask deep NN. We also proposed a novel and more flexible adaptive multitask neural network architecture and showed that it corresponds to a multitask GP with LMC and CC kernels. Our experiments show that the proposed adaptive multitask NN performs better than the standard multitask NN, especially if the tasks are not that highly correlated.

We empirically also show that, for smaller datasets, it is better to use multitask GP than multitask BNN as approximate inference methods like VB that are routinely used for multitask BNN often converge to bad local optima.  We also show that the correlated multitask NN is better than uncorrelated multitask NN.

Exploring the potential relationship between Gaussian Process and more advanced neural network architecture like Transformers~\cite{vaswani2017attention}, multitask BERT~\cite{liu2019multi}, multitask CNN for a better probabilistic understanding of these deep learning models would be a future avenue of this work.

\bibliography{main}{}
\bibliographystyle{plain}

\newpage

\appendix

\section{Multitask Gaussian Process}
In section \textit{Multitask Gaussian Process} we have seen that multitask GP defines a joint Gaussian prior distribution over multiple functions with the following covariance 
\begin{align*}
Cov(f_{t_i}(x_i) , f_{t_j}(x_j)) = K(x_i, t_i, x_j, t_j)
\end{align*}
In this section we derive the predictive posterior distribution for multitask GP and discuss further insights on how information is shared in multitask GP. 
\subsection{Predictive Posterior distribution}
Assuming multitask GP prior on function $f_k$'s  (Gaussian Process prior on joint function $f$) and the following Gaussian Likelihood, 
\begin{align*}
    \mathcal{P}(y | f, x, t) = \mathcal{N}(f(x, t), \sigma^2_{t}) \\
\end{align*}
the predictive posterior distribution for a new data-point $x_*$ and task $t_*$ is be given by,
\begin{align*}
    \mathcal{P}(y_* | x_*, t_*, X, \mathcal{T}, y) &= \mathcal{N}(\mu_*, \sigma_*^2)\text{\quad where}\\
    \mu_* &= K_*^{T}C_N^{-1}y \text{ and } \\
    \sigma_*^2 &= K_{t_*t_*}(x_*,x_*) + \sigma^2_{t_*} - K_*^{T}C_N^{-1} K_* \\
    K_* &= 
    \begin{bmatrix}
        K_{t_*t_1}(x_*, x_1)\\
        K_{t_*t_2}(x_*, x_2)\\
        .\\
        .\\
        K_{t_*t_N}(x_*, x_N )
    \end{bmatrix} \\
    C_N[i, j] &= 
    \begin{cases} 
    K_{t_it_j}(x_i, x_j) + \sigma^2_{t_i} \text{ if } t_i = t_j \\
    K_{t_it_j}(x_i, x_j) \text{ otherwise}
   \end{cases}
\end{align*}
The above equation follows directly from predictive posterior distribution of Gaussian Process regression (since multitask GP prior on $f_1, f_2, .. f_k$ is same as GP prior on joint function $f$).
\subsection{Insights on Information sharing}
Let us examine the the mean of predictive posterior (an Support Vector Machines like interpretation)  
\begin{align*}
    \mu_* &= K_*^{T}C_N^{-1}y  \\
    &= K_*^{T}\alpha \text{ where } \alpha = C_N^{-1}y \\
    &= \sum_{i=1}^{N} K_{t_*t_i}(x_*, x_i)\alpha_i \\
    &= \sum_{i=1}^{N} K_{task}(t_*, t_i)K_{input}(x_*, x_i)\alpha_i \\
\end{align*}
from the above equation we can see that if task $t_i$ is highly correlated to task $t_*$ then data/observation from task $t_i$ contributes more to the prediction, if it is less correlated then it contributes less. 

\subsubsection{Information sharing - Uncorrelated functions}
In this section we show that if the functions are uncorrelated, then multitask GP corresponds to training individual GP for each task. For simplicity lets assume only two tasks, $t_1$ and $t_2$. Let $x_1, x_2, ...x_n \in $ Task $t_1$ and $x_{n+1}, x_{n+2} , ... x_{N} \in $ Task $t_2$ 

Let the task be uncorrelated 
\begin{align*}
    K_{task}(t_i, t_j) = 
    \begin{cases} 
    \omega^2_{t_i} \text{ if } t_i = t_j \\
    0 \text{ otherwise }
    \end{cases} 
\end{align*}
Posterior Mean and variance for input $x$ and task $t_1$ can be written as 
\begin{align*}
   \mu &= K^{T}C_N^{-1}y \text{ and } \\
    \sigma^2 &= K_{t_1t_1}(x,x) + \sigma^2_{t_1} - K^{T}C_N^{-1} K \\
    K &= 
    \begin{bmatrix}
        K_{t_1t_1}(x, x_1)\\
        K_{t_1t_1}(x, x_2)\\
        .\\
        K_{t_1t_1}(x, x_n ) \\
        0 \\
        . \\
        0
    \end{bmatrix} = 
    \begin{bmatrix}
    K^{1} \\
    K^{2}
    \end{bmatrix} \\
    C_N &=  \begin{bmatrix}
    C_{n}^{1} &  0 \\
    0 & C_{N-n}^{2} \\
    \end{bmatrix} \\
    C_N^{-1} &=  \begin{bmatrix}
    (C_{n}^{1})^{-1} &  0 \\
    0 & (C_{N-n}^{2})^{-1} \\
    \end{bmatrix} \\
\end{align*}
Where $C_{n}^{1}$ and $C_{N-n}^{2}$ are covariance matrices corresponding to single task GPs. 
\begin{align*}
    C_n^{1}[i, j] &= 
    \begin{cases}
        K_{t_1t_1}(x_i, x_j) + \sigma^2_{t_1} \text{ if } i=j\\
        K_{t_1t_1}(x_i, x_j) \text{ otherwise}
    \end{cases}
\end{align*}
Now $K^{T}C_N^{-1}$ can be written as 
\begin{align*}
    K^{T}C_N^{-1} &= 
    \begin{bmatrix}
    (K^{1})^{T}(C_{n}^{1})^{-1} &  0
    \end{bmatrix}\\
    \mu &= K^{T}C_N^{-1}y \\
    &= (K^{1})^{T}(C_{n}^{1})^{-1}y^{1} \\
    \text{where }    y &= 
    \begin{bmatrix}
    y_1 \\
    y_2 \\
    \end{bmatrix}\\
    K^{T}C_N^{-1} K &= (K^{1})^{T}(C_{n}^{1})^{-1}(K^{1}) \\
    \sigma^2 &= K_{t_1t_1}(x,x) + \sigma^2_{t_1} - (K^{1})^{T}(C_{n}^{1})^{-1}(K^{1})
\end{align*}
From the above equations we can see that mean and variance of predictive posterior of multitask GP regressions are same as predictive posterior when each tasks are trained individually. Hence if the functions are uncorrelated, then multitask GP corresponds to training individual GP for each task. \\
\textbf{Note: } In multitask GP, hyperparameters can still be shared across multiple tasks (mainly hyperparameters of $K_{input}$ ).

\section{Deep MTNN converges to Deep MTGP }
The derivation for the kernel function for Deep MTGP proceeds recursively. We have seen that input to each hidden layer in Deep MTNN turns out to be a Gaussian i.e. input to $l^{th}$ layer $z^{l-1}(x) \sim GP(0,K^{l-1})$. 
\subsection*{Derivation of multitask GP kernel}
\begin{align*}
    &K^{L(t_1t_2)}(x_1,x_2) = Cov(f_{t_1}(x_1),f_{t_2}(x_2))\\
    &= E[f_{t_1}(x_1)f_{t_2}(x_2)]  \\
    &= E[\big(b^L_{t_1}+\sum_{j=1}^{N_L}v_j^{t_1}x^L(x_1)\big)\big(b^L_{t_2}+\sum_{k=1}^{N_L}v_k^{t_2}x^L(x_2)\big)]\\
    &= E[b^L_{t_1}b^L_{t_2}+\sum_{j=1}^{N_L}\sum_{k=1}^{N_L}v_j^{t_1}v_k^{t_2}x^L(x_1)x^L(x_2)]\\
    &= \sigma^{2(t_1t_2)}_b + \sum_{j=1}^{N_L} E[v_j^{t_1}v_j^{t_2}]E[x^L(x_1)x^L(x_2)]\\
    &= \sigma^{2(t_1t_2)}_b +\sum_{j=1}^{N_L} \sigma_v^{2(t_1t_2)}E[h^L(z_j^{L-1}(x_1))h^L(z_j^{L-1}(x_2))]\\
    &= \sigma^{2(t_1t_2)}_b +\\ &\sum_{j=1}^{N_L} \sigma_v^{2(t_1t_2)}E_{z_j^{L-1} \sim GP(0,K^{L-1})}[h^L(z_j^{L-1}(x_1))h^L(z_j^{L-1}(x_2))]\\
    &= \sigma^{2(t_1t_2)}_b +\\ &\sigma_v^{2(t_1t_2)}N_L E_{z^{L-1} \sim GP(0,K^{L-1})}[h^L(z^{L-1}(x_1))h^L(z^{L-1}(x_2))]\\
    &= \sigma^{2(t_1t_2)}_b +\\ &\omega_v^{2(t_1t_2)} E_{z^{L-1} \sim GP(0,K^{L-1})}[h^L(z^{L-1}(x_1))h^L(z^{L-1}(x_2))]\\
    &= \sigma^{2(t_1t_2)}_b +\\ &\omega_v^{2(t_1t_2)} F_h(K^{L-1}(x_1,x_1), K^{L-1}(x_1,x_2), K^{L-1}(x_2,x_2))\\
\end{align*}


\section{Limitations of ICM Kernels}

We have seen that hard-parameter multitask neural network corresponds to Gaussian Process with the well known Intrinsic coregionalization model (ICM) kernels\cite{alvarez2012kernels}. Despite its simplicity and popularity its a very rigid form of kernels, one of the major drawbacks of ICM kernels is that it does not give any guarantee that learning multiple task together performs atleast as good as learning each task individually (assuming we can find global optima of parameters and hyperparameters). This restricts the power of multitask learning and requires humans(domain knowledge) to carefully choose the related tasks to improve performance using MTL. The reason for this rigidity is the single $K_{input}(x_1,x_2)$ and hence the GPs can model only one type of function, i.e. it assumes all the tasks to have similar properties (like smoothness, differentiability, periodicity etc) and this could make MTL (both above mentioned MTGP and MTNN) perform worse than training each task independently if the tasks have different properties. 

However, a true MTGP formulation does not make any assumptions on task relatedness, i.e., MTGP does not assume tasks to have similar properties. Therefore, the limitation is due to the kernel choice and not with the theory of Multitask Gaussian Process. 
\begin{align*}
     Cov(f(x_i, t_i) , f(x_j, t_j))  &= K(x_i, t_i ,x_j, t_j) 
\end{align*} 
Although not as general as above kernels (i.e., a separate kernel function for each pair of tasks), Linear model of coregionalization (LMC) and Cross Coregionalization kernels \cite{alvarez2012kernels} are much more flexible than ICM kernels.

Linear Model of Coregionalization(LMC) kernel
\begin{align*}
   K(x_1,t_1,x_2,t_2) = \sum_{m=1}^{\mathcal{K}} K^m_{task}(t_1,t_2) K^m_{input}(x_1,x_2) 
\end{align*}

Cross-coregionalization kernel
\begin{align*}
    K(\bm{x_1},t_1,\bm{x_2},t_2) = \sum_{m=1}^{\mathcal{K}}\sum_{n=1}^{\mathcal{K}} K^{m,n}_{task}(t_1,t_2) K^{m,n}_{input}(x_1,x_2) \\
\end{align*}

We can see that for $\mathcal{K} = T$(number of tasks) LMC kernels can perform atleast as good as learning each task individually (and hence Cross-coregionalization kernel), i.e. 
\begin{align*}
   K(x_1,t_1,x_2,t_2) &= \sum_{m=1}^{\mathcal{T}} K^m_{task}(t_1,t_2) K^m_{input}(x_1,x_2)\\
   &= K^{t}_{input}(x_1,x_2) \text{ if } t_1 = t_2 = t\\
   &= 0 \text{ otherwise}
\end{align*}
where $K^{t}_{input}(x_1,x_2)$ is the kernel of task $t$ when training individually. We can also see that for $\mathcal{K} = T$, Cross-coregionalization kernels can form the general MTGP kernels (a separate kernel for each pair of tasks).

\section{Proof: Adaptive MTBNN Converges to Multitask GP}
We can write the output of task $t$, $f_{t}(x)$ as 
\begin{align*}
    f_t(x) &= \sum_{k=1}^{\mathcal{K}} \sum_{j=1}^{H} v^{k}_{jt} h^{k}(x, U_j^{k}) \\
    \text{Let } z^{k}_{jt} &=\sqrt{H} v^{k}_{jt} 
\end{align*}
We will show that for any finite subset of input $x_1, x_2, ... x_s$ with corresponding task $t_1, t_2, .. t_s$, the output of adaptive multitask NN jointly converges to a Gaussian distribution.
\begin{align*}
    \begin{bmatrix}
        f_{t_1}(x_1)\\
        f_{t_2}(x_2)\\
        .\\
        .\\
        f_{t_s}(x_s)
    \end{bmatrix} &= \sqrt{H}\Big[ \frac{1}{H}\sum_{j=1}^{\mathcal{H}} \Big(
    \sum_{k=1}^{K}  F_{jk} 
        \Big) \Big]\\
    \text{Where } F_{jk} &= \begin{bmatrix}
            z^{k}_{jt_1} h^{k}(x_1, U_j^{k}) \\
            z^{k}_{jt_2} h^{k}(x_2, U_j^{k})\\
            .\\
            .\\
            z^{k}_{jt_s} h^{k}(x_s, U_j^{k})
        \end{bmatrix} \\
    \text{Let } G_j &= \sum_{k=1}^{K}  F_{jk} 
\end{align*}

Note that $F_{ik}$ and $F_{jk}$ are identical for any $i, j$ (as $ z^{k}_{it}$, $z^{k}_{jt}$  are identical and $U_i^{k}$, $U_j^{k}$ are identical).Hence $ G_j = \sum_{k=1}^{K}  F_{jk} $ are identical for all $j$. 

Also note that $F_{ik_a}$ and $F_{jk_b}$ are independent for all $k_a, k_b, i\neq j$, as $z^{k_a}_{it}$, $z^{k_b}_{jt}$ and  $U_i^{k_a}$, $U_j^{k_b}$ are also independent for all $k_a, k_b, i \neq j$. Hence $G_i$ and $G_j$ are independent (every element/coordinate in $G_i$ is independent of $G_j$). As $F_{ik}$ has a finite variance (refer normal multitask NN), $G_j$ also has finite variance. Hence by applying multidimensional Central Limit theorem, $\sqrt{H}\Big[ \frac{1}{H}\sum_{j=1}^{\mathcal{H}} G_j \Big]$ converges to Gaussian distribution. Hence we show that adaptive multitask NN jointly converges to a multitask GP. 

\section{Experiments}
\subsection{Dataset Descriptions:}
In this section we briefly describe each dataset used in our experiments. 
\subsubsection{SARCOS: }
SARCOS data~\cite{williams2006gaussian} is a mapping from 7 positions, 7 velocities, 7 accelerations to the corresponding 7 joint torques of SARCOS anthropomorphic robot arm.  
We only used first 600 data from training set and first 500 data from test set,(In our experiments we use exact GP inference which is not scalable for large data size, hence we sampled a small subset of data). Also, we used only used first and second outputs (2 torques) to conduct our experiments (as increasing the number of tasks increases number of hyperparameters and effective size of data).

\subsubsection{Polymer Dataset} The input of this dataset consists of 10 controlled variables of a polymer processing plant that are mapped to 4 target variables which are the measurements of the output of that plant. Again we used only first 2 outputs for our experiments. 

\subsubsection{SICK: }
Sentences Involving Compositional Knowledge  (SICK) is a small textual Entailment dataset, each data point consist of two sentences (Premise and Hypothesis) and their label (Entailment, Neutral and Contradiction). It is a part of SemEval-2014, we use SICK entailment (SICK-E) data for our experiments.  
\subsubsection{MultiNLI: }
Multi-Genre Natural Language Inference (MultiNLI) is a huge crowd-sourced textual entailment dataset (about 400k sentence pairs). It covers different genres like fiction, letter, telephone Speech, 9/11 Report etc.. 

\subsubsection{SST-2: }
Stanford Sentiment Treebank - Binary classification (SST-2) contains binary sentiment labels (positive and negative) for each sentence. It contains about $60k$ examples.


\subsection{Advantage of Adaptive Multitask NN - Simulated dataset}
We simulate two tasks carefully such that they possess different properties and show that hard-parameter multitask neural networks (MTGP with ICM kernels) may perform worse than training each task independently, and in such cases adaptive multitask neural network (MTGP with LMC kernels) could perform better. 

\subsubsection{Simulated Dataset:}
We sampled 2 independent functions from Gaussian Process prior with logistic kernels\cite{williams1997computing} but with significantly different hyperparameters (i.e. different properties).
\begin{figure}[!htbp]
    \centering
    \includegraphics[scale=0.4]{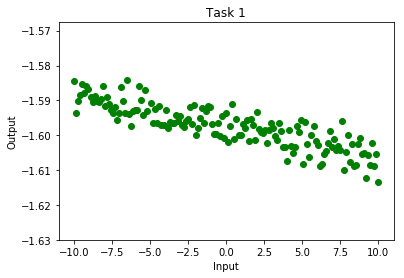}
    \caption{Task 1 - Gaussian Process sample}
    \label{fig:adap_task1}
\end{figure}
\begin{figure}[!htbp]
    \centering
    \includegraphics[scale=0.4]{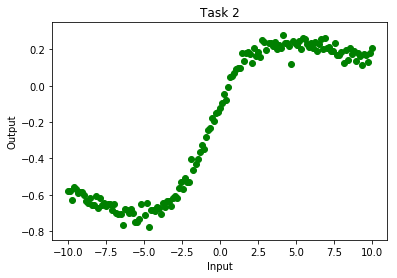}
    \caption{Task 2 - Gaussian Process sample}
    \label{fig:adap_task2}
\end{figure}

Logistic Kernel: 
\begin{align*}
    K(x_1, x_2) &= c + \omega^2\frac{2}{\pi} sin^{-1} \frac{2x_1^T\Sigma x_2}{\sqrt{(1 + 2x_1^T\Sigma x_1)(2x_2^T\Sigma x_2) }}
\end{align*}
while sampling we fixed $\Sigma = 10^{k} * I_2 $, we used different $k$'s for each of the task. We also observed that as the $k$ value increases the function becomes more rough/random (and for smaller $k$'s functions are more smoother). Since both the task has different properties a single $K_{input}$ function may not work properly. 

\subsubsection{Results:}
We report mean squared errors (MSE) on both task1 and task2. In this data,  results of task 1 and task 2 are significantly different (approximately 10 times) therefore for multitask learning minimising average MSE is not appropriate, hence we minimized weighted average of MSE of both the tasks, where weights are chosen using validation error of individual neural networks.  

\begin{table}[!htbp]
\centering
\caption{MTNN vs Adaptive MTNN}
\begin{tabular}{c  c c c c} 
\toprule
& \textbf{single task NN} & \textbf{MTNN} & \textbf{Adap. MTNN} \\
\midrule
\multicolumn{4}{c}{Task 1} \\
\midrule
MSE dev &0.29  &  0.50 &0.38 \\
MSE Test & 0.23 & 0.46 & 0.33 \\
\midrule
\multicolumn{4}{c}{Task 2} \\
\midrule
MSE dev & 0.029 &  0.048 & 0.069 \\
MSE Test &0.040  &  0.060 & 0.026  \\
\bottomrule
\end{tabular}
\label{tbl:mtbnn-vs-adap}
\end{table}

\end{document}